\documentclass[letterpaper, 10 pt, conference]{ieeeconf}  % Comment this line out if you need a4paper

\IEEEoverridecommandlockouts                              % This command is only needed if 
\usepackage{epsfig} % for postscript graphics files
\usepackage{amsmath} % assumes amsmath package installed
\usepackage{multirow}
\usepackage{booktabs}
\usepackage{fancyhdr}

\DeclareMathAlphabet{\mathbb}{U}{bbold}{m}{n}
\title{\LARGE \bf
Learning State-Space Models for Mapping Spatial Motion Patterns
}

\author{Junyi Shi$^{1}$ and Tomasz Piotr Kucner$^{1,2}$% <-this % stops a space
\thanks{$^{1}$Junyi Shi and Tomasz Piotr Kucner are with the Department of Electrical Engineering and Automation, Aalto University, Finland.
        {\tt\small junyi.shi, tomasz.kucner@aalto.fi}}%
\thanks{$^{2}$Tomasz Piotr Kuncer is also with the Finnish Center of Artificial Intelligence, Finland.
        }%
}

\begin{document}

\maketitle
\thispagestyle{empty}
\pagestyle{empty}

\thispagestyle{fancy}
\fancyhead{}
\lhead{}
\lfoot{979--8-3503-0704-7/23/\$31.00 \copyright2023 IEEE}
\cfoot{}
\rfoot{}

%%%%%%%%%%%%%%%%%%%%%%%%%%%%%%%%%%%%%%%%%%%%%%%%%%%%%%%%%%%%%%%%%%%%%%%%%%%%%%%%
\begin{abstract}

Mapping the surrounding environment is essential for the successful operation of autonomous robots. While extensive research has focused on mapping geometric structures and static objects, the environment is also influenced by the movement of dynamic objects. Incorporating information about spatial motion patterns can allow mobile robots to navigate and operate successfully in populated areas. In this paper, we propose a deep state-space model that learns the map representations of spatial motion patterns and how they change over time at a certain place. To evaluate our methods, we use two different datasets: one generated dataset with specific motion patterns and another with real-world pedestrian data. We test the performance of our model by evaluating its learning ability, mapping quality, and application to downstream tasks. The results demonstrate that our model can effectively learn the corresponding motion pattern, and has the potential to be applied to robotic application tasks.

\end{abstract}

%%%%%%%%%%%%%%%%%%%%%%%%%%%%%%%%%%%%%%%%%%%%%%%%%%%%%%%%%%%%%%%%%%%%%%%%%%%%%%%%
\section{Introduction}
% In recent years, mobile robots have been gradually applied to a large number of scenarios such as indoor logistics, field exploration, and aerial flight, etc. Map, as a common way of modelling environmental information, can benefit robotics applications in many ways. However, mobile robots still suffer from their limited capabilities in changing environments. To allow mobile robots successfully navigate and operate in a populated area, we need to implement methods to map the information of the dynamics. 

% In recent years, mobile robots have found extensive applications in diverse scenarios, such as indoor logistics, healthcare and exploration. The map is a widely adopted approach for modeling environmental information and can provide several benefits for robotics applications. However, mobile robots encounter limitations in dealing with changing environments. To allow mobile robots successfully navigate and operate in populated areas, Thus, methods need to be implemented to map the dynamics' information, enabling mobile robots to navigate and operate successfully in populated areas.

% In recent years, mobile robots have been increasingly utilized in a wide range of scenarios, including logistics,  healthcare and exploration, etc. Map, as a common way of modelling environmental information, can benefit robotics applications in many ways. However, mobile robots still encounter limitations in dealing with changing environments. To allow mobile robots successfully navigate and operate in populated areas, it is necessary to develop methods for mapping dynamic information.

In recent years,  the utilization of mobile robots has witnessed significant growth across various applications such as logistics, healthcare and exploration. Mapping, serving as a fundamental approach for modeling environmental information, plays a vital role in enabling robots to plan their movements, avoid obstacles, and locate targets. However, mobile robots still encounter limitations in dealing with changing environments. To allow mobile robots successfully navigate and operate in populated areas, it is necessary to develop methods for mapping dynamic information.

In daily life, it can be observed that individuals often adhere to implicit traffic rules while navigating their surroundings. Pedestrians exhibit distinct movement depending on their location, such as when traversing a corridor or approaching building entrances. Moreover, people from different regions tend to follow specific directional norms. For instance, individuals in the UK and Japan tend to favor the left side, while those in the US and Canada exhibit different behaviors. This observation naturally gives rise to a hypothesis: there exists a spatial motion pattern that guides the movement of pedestrians. With the map representation of these motion patterns, mobiles robot can benefit in a variety of applications such as motion planning \cite{swaminathan2022benchmarking}, human motion prediction \cite{humanmotionprediction}, task planning \cite{taskplanning}, and human-robot interaction \cite{humanrobotinteraction}.

% Modelling these spatial motion patterns can be challenging and time-consuming. Previous works are either based on the assumption of motion pattern won't change in a given location \cite{CLiFF}, or changes severely over a long period of time \cite{Fremen}. H1owever, these assumptions are to some extent divorced from reality. In reality, the pattern of movement is gradually changing: at an underground station, the number of people neither remains the same nor increases instantaneously, but changes slightly when gradually approaching a certain rush hour.

Modelling these spatial motion patterns can be challenging. Previous studies are either based on the assumption that motion patterns remain constant within a given location \cite{CLiFF} or undergo significant changes over extended periods \cite{Fremen}. However, such assumptions are somewhat divorced from reality. In reality, motion patterns tend to evolve gradually, as seen in an example of an underground station where the number of people does not remain constant, nor does it increase instantaneously. Instead, it changes gradually as the station approaches a certain rush hour.

In this paper, we adopt the assumption that dynamics within a changeable environment  are driven by a certain kind of motion pattern that undergoes gradual changes over time. Our approach focuses on learning a map representation that describes the implicit motion pattern and its temporal variations. By leveraging data collected over successive time periods, our method can effectively learn the corresponding motion patterns and predict their subsequent movements.

% In this paper, we assume that the dynamics in a changeable environment are driven by a certain kind of motion pattern, which slightly changes over time. We learned a map representation to describe the implicit motion pattern and also its changes. As shown in Fig. \ref{vortexshow}, it describes the dynamics observed in a toy vortex dataset, which exhibits a distinct "vortex" movement and rotates over time. With data collected over successive time periods, our method can effectively learn the corresponding motion pattern and predict their subsequent movements.

Our contributions can be summarised as follows:
\begin{itemize}
    \item We implement a generative model to describe the spatial motion pattern, which aggregates and encodes the spatial information of dynamics into a map representation.
    \item We employ a state-space model (SSM) to represent how the spatial motion pattern changes over time at a certain place.
    \item We demonstrate the predictive performance using our learned model, by evaluating the learning ability, the mapping quality and the model’s applicability to downstream tasks.
\end{itemize}

\section{Related Work}

% \subsection{Maps of Dynamics}
Our work is based on the concept of \textit{Maps of Dynamics} (MoD),  which refers to spatial or spatio-temporal representations of patterns of dynamics \cite{KucnerBook}. 
 \begin{figure*}[htbp]
      \centering
      \includegraphics[width=12cm]{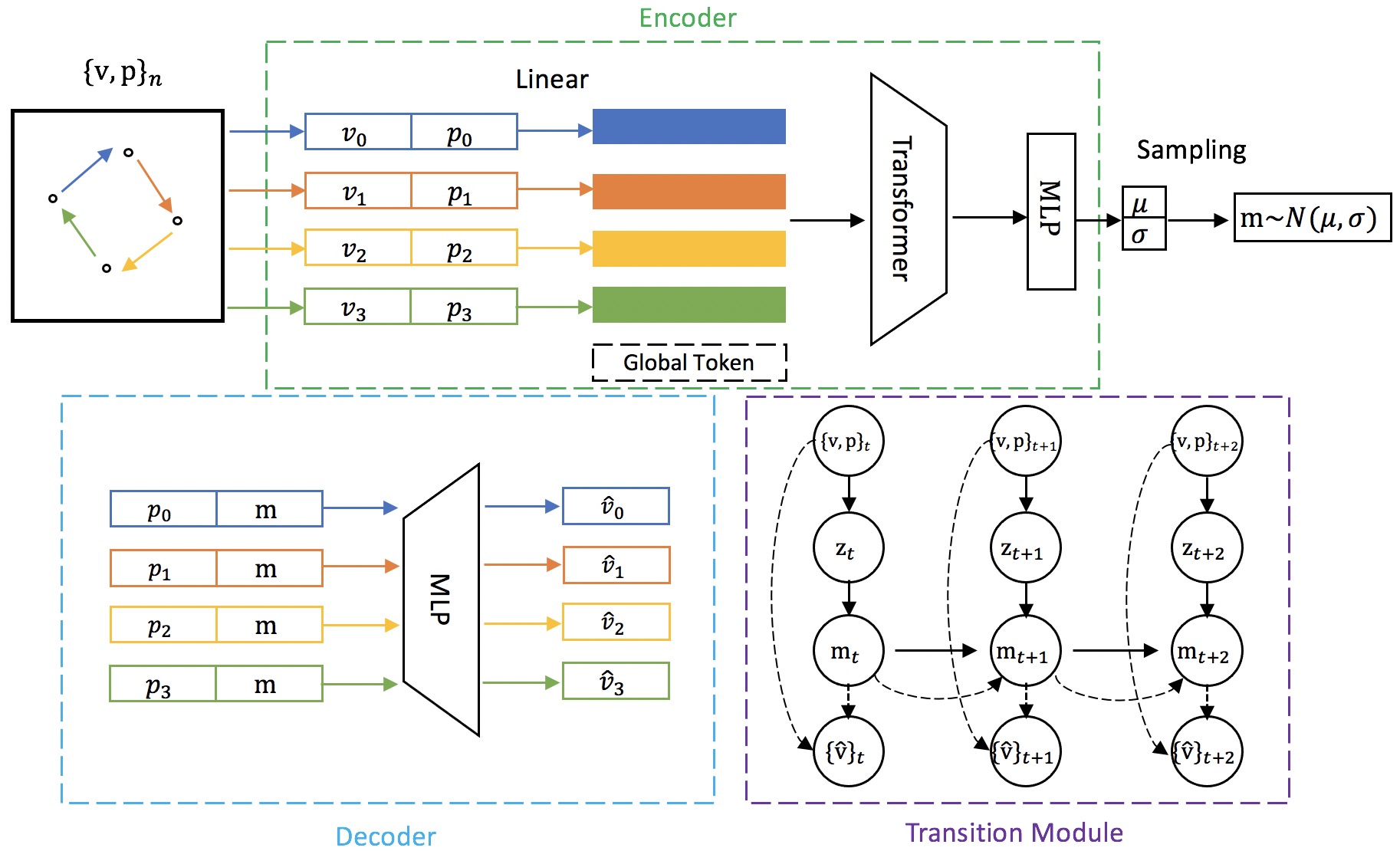}
      \caption{Overview of our system. Our model consists of three major components: Encoder, Decoder and Transition Module. The encoder takes the pairs of velocity $\mathbf{v}$ and position $\mathbf{p}$ as input, where the Transformer converts a set of points to a single feature vector $\mathbf{z}$ and output a normal distribution. After sampling, the system obtains the motion pattern $\mathbf{m}$. The decoder takes position $\mathbf{p}$ and motion pattern $\mathbf{m}$ as inputs, and generate the reconstruction $\mathbf{\hat{v}}$. Observations from the past state are used to generate predictions into the future state in the transition module. }
      \label{Model}
  \end{figure*}

MoDs can be classified into different groups based on the type of dynamics being mapped. When considering discrete objects, they can be classified into three main groups: static objects, semi-static objects, and dynamic objects.  \cite{meyer2010temporary}.  Static objects, such as trees and buildings, rarely change position over long periods of time. In mapping systems, these are often represented using geometric maps, such as occupancy grid map \cite{grid} or OctoMap \cite{octomap}, which are not considered as MoDs. Semi-static objects, such as chairs and boxes, might change position within a relatively low frequency or as a consequence of specific events. Krajník et al. \cite{Fremen} introduce occupancy grids for mapping semi-static objects, combined with the temporal model Frequency Map Enhancement (FreMEn), in order to model the state changes of the semi-static cells. Dynamic objects, such as pedestrians and animals, are some objects that move purposefully and can be observed during the change of their states. Kucner et al. \cite{CTMap} and Wang et al. \cite{IOHMM} treat dynamics as a change of occupancy in grid map cells and construct models capable of grasping the spatial relation between the states of neighboring cells. Dynamic objects can also be modelled by their trajectory, as explored by Bennewitz et al. \cite{bennewitz2002learning} and Ellis et al. \cite{ellis2009modelling}, or represented by velocity fields, as proposed by  Verdoja et al. \cite{verdoja2022generating} and CLiFF-Map \cite{CLiFF}.

Traditionally, state-space models have been used to produce estimates of currently unknown state variables based on their previous observations \cite{sarkka2013bayesian}. As a common approach, it is widely used in applications such as state estimation \cite{sarkka2007unscented}, target tracking \cite{tracking} and navigation \cite{grewal2007global}. In recent years, deep sequential generative models are appealing as temporal models, which have shown impressive performance in various types of inference tasks, such as system identification \cite{gedon2021deep}, geometric mapping \cite{mirchev2020variational}. By learning from past experience, it can be applied to model environmental dynamics and uncertainty due to the probabilistic nature of the model.

In this paper, we focus primarily on mapping spatial motion patterns of dynamic entities. In contrast to previous studies, we adopt the assumption that the motion patterns of these entities undergo gradual changes within short time frames, and can be implicitly represented. We propose a deep sequential generative model specifically designed for the MoD problem, by learning a state-space model that represents the underlying motion patterns based on past experiences.

% ***

% Grid-based, discrete work

% continous work

% no changeable, Fremen - severly changed in 3 hours 

% ***

% \subsection{Generative Models}
% VAE, GAN, Diffusion Model

% ***

% We use VAE, because it's not in given encoder forms of GAN/Diffusion 

% ***

% \subsection{State-Space Models}
% Kalman Filter -- learning models

% ***

\section{Methodology}
\subsection{Problem Setup}

In our work,  we employ a motion probability distribution to represent the dynamics. The motion distribution, denoted as $\mathcal{M}$, is defined as a conditional distribution of velocity $\mathbf{v}$ given position $\mathbf{p}$:
\begin{equation}
    \mathcal{M} = p(\mathbf{v} \mid \mathbf{p}),
\end{equation}
where $\mathbf{p}$ is a 2D Euclidean vector denoting the position. The velocity $\mathbf{v}$ is using a polar coordinate frame, which combines the orientation $\psi$ and speed $\rho$:
\begin{equation}
\mathbf{v}=(\psi, \rho)^{\top}, \psi \in[-\pi, \pi)
\label{velocity}
\end{equation}
By using a polar representation rather than a 2D Euclidean vector representation, each component of the velocity vector has an explicit physical meaning and can be analyzed independently.

At each time step $t$ we are interested in, we make the assumption that there are no changes in the underlying motion pattern. We observe $n$ points at the time step $t$ in the form of $\{\mathbf{v}, \mathbf{p}\}$, which can be viewed as samples from the joint distribution:
\begin{equation}
   p(\mathbf{v}_t, \mathbf{p}_t) = p(\mathbf{p_t}) p(\mathbf{v}_t \mid \mathbf{p}_t)
    \label{MotionDistribution}
\end{equation}

We further assume that the dynamics in the given location are driven by a spatial motion pattern $\mathbf{m}$, which serves as a parameter of the motion distribution. Different values of $\mathbf{m}$ give rise to distinct motion distributions. 
The joint distribution with $\mathbf{m}$, conditioned upon Equation (\ref{MotionDistribution}), can be expressed as follows:

% Thus, we can extend Equation (\ref{MotionDistribution}) as follows:
\begin{equation}
    p(\mathbf{v}_t, \mathbf{p}_t \mid \mathbf{m}) = p(\mathbf{p}_t) p(\mathbf{v}_t \mid \mathbf{p}_t, \mathbf{m}), 
\end{equation}
where we assume the position $\mathbf{p}$ is independent of $\mathbf{m}$.

The purpose of this formulation is to estimate the motion distribution based on the given set of observations. In practical implementation, we utilize Gaussian distributions for all the distributions in our formulation. Specifically, we employ a neural network that outputs both the mean and variance of the Gaussian distribution, allowing us to learn and estimate the parameters of the motion distribution.

\subsection{Network Structure}
In our work, we employ variational inference and amortized inference \cite{kingma2013auto} techniques to address the problem. The neural network utilized in our approach consists of three distinct components: the approximate posterior distribution defined as $q_\phi(\mathbf{m} \mid \{\mathbf{v}, \mathbf{p}\}_n)$, the prior distribution $p_\theta(\mathbf{m})$ and the emission model $p_\theta(\mathbf{v} \mid \mathbf{p}, \mathbf{m})$.

To handle the varying number of observations at each time step, we introduce the concept of a set feature extractor. The set feature extractor converts a set of points to a single feature vector: $\mathbf{z} = f(\{\mathbf{v}, \mathbf{p}\}_n)$. The set feature extractor allows us to align the encoder with other components and simplify the dependencies on the motion set $\{\mathbf{v}, \mathbf{p}\}_n$.

Based on this, the posterior is split into two parts. First, a set feature extractor is applied to convert the set into a vector representation. Then, Multilayer Perceptrons (MLPs) are applied to compute the mean and variance of the posterior distribution. There are multiple options available for the set feature extractor, we choose Transformer \cite{vaswani2017attention} as the extractor for its reliable performance. 

In most cases, the variational autoencoder (VAE) does not require the learning of the prior distribution, a standard Gaussian $\mathcal{N}$ can be simply utilized. Furthermore, a  specialized decoder can only be implemented with a known state structure, such as the motion pattern $\mathbf{m}$ in our case. Therefore, we utilize a common flatten decoder, which is a combination of several MLPs.

We employ \textit{evidence lower bound} (ELBO) \cite{VariationalMethods} as the objective function, which is given as:
\begin{equation}
\begin{aligned}
\mathcal{L}_{elbo} = & \mathbb{E}_{\{\mathbf{v}, \mathbf{p}\}_n \sim D} \Big[\mathbb{E}_{\mathbf{m} \sim q_\phi(\mathbf{m} | \{\mathbf{v}, \mathbf{p}\}_n)}\\
& [\frac{1}{n} \sum_{i=1}^n - \log p_\theta(\mathbf{v}_i | \mathbf{p}_i, \mathbf{m})] + \\
& D_{KL}[q_\phi(\mathbf{m} | \{\mathbf{v}, \mathbf{p}\}_n) || p_\theta(\mathbf{m})] \Big],
\end{aligned}
\end{equation}
The ELBO provides a lower bound on the marginal likelihood, which is intractable to compute directly. Maximizing the ELBO is equivalent to minimizing the Kullback-Leibler (KL) divergence between the approximated posterior and the true posterior. 

% The ELBO is a commonly used objective function in variational inference and has been shown to be effective in many applications.

\subsection{Sequential Modelling}

Since we assume that there is a underling law that guiding the changes of motion pattern, we can extend our model to handle sequential data using the state-space model formulation.

To accomplish this, we extend the posterior and the prior distributions to a sequential form, which can be expressed as follows:
\begin{equation}
    \begin{array}{lll}
       \mbox{Posterior: } & \mathbf{m}_{t+1} & \sim q_{\phi}\left(\mathbf{m}_{t+1} \mid \mathbf{m}_{t},\{\mathbf{v}, \mathbf{p}\}_{n_{t+1}}\right)   \\
       \mbox{Prior: } & \mathbf{m}_{t+1} & \sim p_\theta\left(\mathbf{m}_{t+1} \mid \mathbf{m}_{t}\right) 
    \end{array}
\end{equation}

In the sequential modelling, the decoder remains the same as the VAE model. Specifically, we employ a recurrent state-space model (RSSM) proposed by Hafner et. al \cite{RSSM}, which is one of the state-of-the-art SSMs. Typically, transitions in a recurrent neural network are purely deterministic, while transitions in a state-space model are purely stochastic. RSSM uses a mix of deterministic and stochastic latent state, which allow the model itself to robustly learn to predict multiple future states. For the SSM, we also utilize MLPs to compute the mean and variance. The whole structure of the model is shown in Figure \ref{Model}.

\section{Experiments}

% Toy data 

% ***

% Real-world scenes, downstream applications

We evaluated three aspects of our MoDs: the learning ability, the mapping quality and the model's potential applicability to downstream tasks. 

The model described in Section III is implemented in PyTorch  \cite{paszke2019pytorch}. We employed GRU \cite{GRU} as the recurrent neural network (RNN) in our model for the deterministic transition. The dimension of the set feature extractor is 256, the dimension of the hidden state for encoder is 1024 and decoder is 256, the dimension for the deterministic transition is 512, the dimension for the stochastic transition is 256 and the dimension of the latent variable is 256.

\subsection{Evaluation of Learning Ability}
\begin{figure}[htbp]
      \centering
      \includegraphics[width=7.5cm]{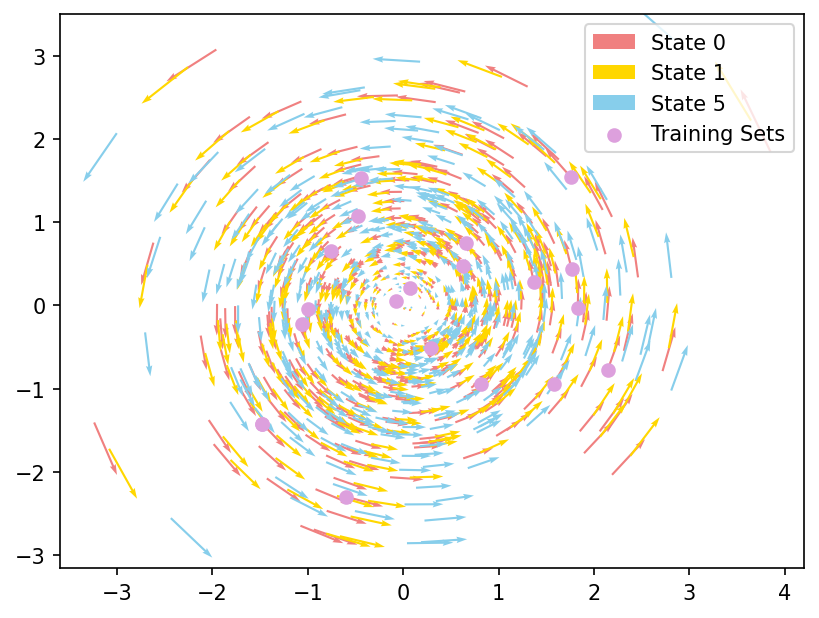}
      \caption{The toy vortex dataset.}
      \label{toy}
  \end{figure}

We started our evaluation with a generated toy dataset, which has clear, explicit motion patterns. A vortex-like pattern is defined as:
\begin{align}
    \dot \rho &= 0.5 \rho, \\
    \dot \psi &= \psi + \frac{\pi}{2}.
\end{align}
The velocity fields of the vortex pattern are generated using \texttt{scipy.integrate.odeint} \cite{2020SciPy-NMeth}, as shown in Fig. \ref{toy}.
  
We trained the model with a batch size of 16 in 500 epochs using AdamW \cite{loshchilovdecoupled} with an learning rate of 0.001. In order to simulate a realistic scenario, only 20 randomly selected velocities in every time step were used as the training set. 20 time steps were used for training, the model observed 5 time steps and predicted 5 time steps or 20 time steps in the experiment.

\begin{figure}[htbp]
      \centering
      \includegraphics[width=8cm]{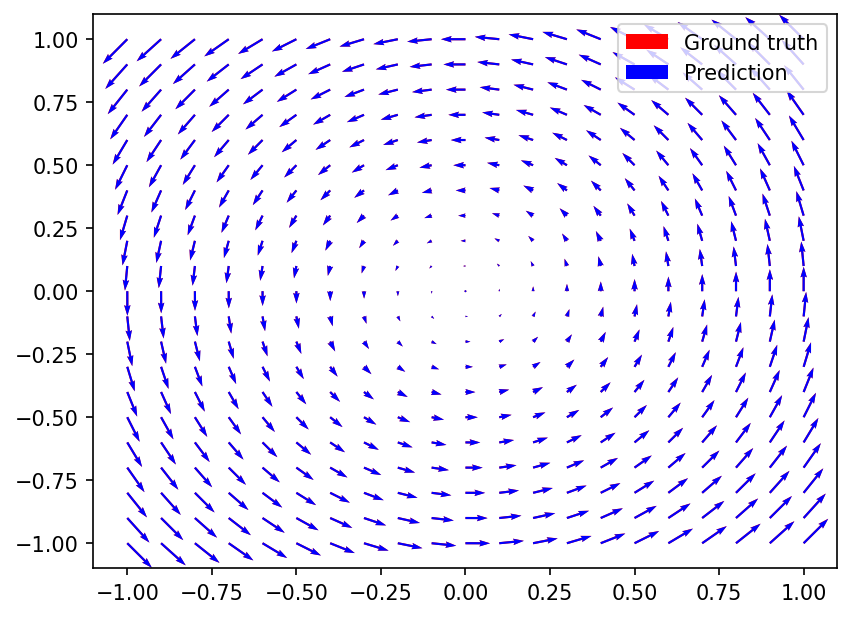}
      \caption{Results of a vortex-like spatial motion pattern. The output of the proposed method is shown in blue, which overlaid with the ground truth in red. The arrows indicate the magnitude and orientation of the velocity field at this future time. }
      \label{vortexshow}
  \end{figure}

As shown in Fig. \ref{vortexshow}, tests were done on another generated vortex dataset, which demonstrated the performance of our model in a similar scenario. We analysed the velocity predicted by our model, considering the magnitude of the velocity error at the ground-truth location. Two metrics were used: Average Velocity Error (AVE) and Final Velocity Error (FVE). The former metric is calculated by the mean square error (MSE) over all estimated points of the states and the true points, while the latter one is calculated at the predicted final state. 

The results are shown in Table \ref{vortex_eva}, where we tested the errors from the SSM compared to the baseline VAE. SSM demonstrated a strong learning capability, performing well in both AVE and FVE metrics. As the other parameters of the two networks are identical, SSM only adds the transition module, so it can be assumed that this increases the ability of our model to learn changes in motion patterns over time.
\begin{table}[htbp]
    \centering
    \caption{Experimental results on vortex dataset}
    \begin{tabular}{lccc}
    \toprule
    Model  &Horizon &  AVE & FVE \\
    \midrule
    \multirow{2}{*}{VAE} & 5 time step & 0.00591 & 0.00595 \\
    & 20 time step &  0.00601 & 0.00584\\
    \multirow{2}{*}{SSM} & 5 time step & 0.00004 & \textbf{0.00006} \\
    & 20 time step & \textbf{0.00003} & 0.00007\\
    \bottomrule
    \end{tabular}
    
    \label{vortex_eva}
\end{table}

\subsection{Quantitative Evaluation}
        Experimenting in a simulated environment was not enough, so we further introduced real-world datasets for evaluation. However, in real scenarios, we are unable to obtain true values of the motion patterns.  Therefore, we implemented the quantitative evaluation to assess the mapping quality of the representation.

\begin{figure}[htbp]
      \centering
      \includegraphics[width=8cm]{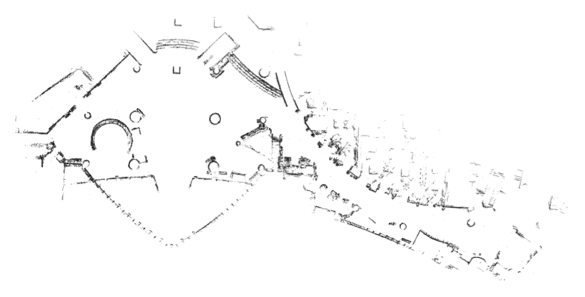}
      \caption{The occupancy grid map of the ATC shopping mall.}
      \label{grid}
  \end{figure}

The ATC dataset \cite{brvsvcic2013person} was used in the experiments, which comprised real pedestrian data from the Asia and Pacific Trade Center in Osaka, Japan. The dataset was obtained using a tracking system comprising numerous 3D range sensors, covering an area about 900 $m^2$. The data collection took place over 92 days between 24 October 2012 and 29 November 2013, specifically on Wednesdays and Sundays, between the hours of 9:40 and 20:20. The spatial geometric map for the environment is shown in Figure \ref{grid}, which contains a long corridor and several entrances.

A divergence estimator proposed by Wang et al. \cite{wang2009divergence} was used to provide the quality of the map in absolute values. Wang's divergence estimator computes the differences between the output of the model and the original data. For each query location in the map, we obtain a motion distribution from the output of the encoder. Simultaneously, we also have a set of observations $\left\{\mathbf{v}_1^o, \ldots, \mathbf{v}_n^o\right\}$ from the given dataset. Wang's divergence estimator was then employed to estimate the divergence between the above two distributions, which employed only the samples coming from them. The estimator is given as follows:
\begin{equation}
\hat{D}_{n, m}\left(\mathcal{M}^{'} \| \mathcal{M}\right)=\frac{d}{n} \sum_{i=1}^n \log _2 \frac{v_k(i)}{\rho_k(i)}+\log _2 \frac{m}{n-1}
\end{equation}

In the divergence estimation, the distance $\rho_k(i)$ between $\mathbf{v}_i^o$ and its k-NN in $\left\{\mathbf{v}_j^o\right\}_{j \neq i}$ is compared with the distance $v_k(i)$ between $\mathbf{v}_i^o$ and its k-NN in $\left\{\mathbf{v}_j^q\right\}$, where $\left\{\mathbf{v}_j^q\right\}$ denotes the observations queried from the component of the model.
 
We retrained the model with a batch size of 16 in 1000 epochs, 20 days are used for training, 5 for validation, a Sunday set and a Wednesday set for evaluation. We take half an hour as a time step, and divide the day into 20 time steps. We employed CLiFF-Map \cite{CLiFF} as a baseline method for comparison, which were trained in different time steps. We set k=1 in practice, which means the algorithm only considers the closest single neighbor to the new data point.

% represented by 11AM and 12AM, and calculate the average.

The result of the quantitative evaluation is shown in Table \ref{Divergence}.  On Sundays, the observations are roughly twice as high as on Wednesdays and the time period starting at 12AM is usually the peak of crowd density. Both methods are influenced by changes in observations, and our method is less sensitive to population density than CLiFF-Map. Since this experiment is actually comparing the ability to aggregate information (either through clustering in CLiFF-Map or through the set feature extractor in our method), it can be shown that our method is more robust to the number of observations.

\begin{table}[htbp]
    \centering
    \caption{Quantitative evaluation results}
    \begin{tabular}{lccccc}
    \toprule
   \multirow{2}{*}{Model} & \multirow{2}{*}{Horizon} &  \multicolumn{2}{c}{Sun} & \multicolumn{2}{c}{Wed}\\
   
     &  &Obs.& Div.[bit] &Obs.& Div.[bit]\\
    \midrule
    \multirow{3}{*}{CLiFF \cite{CLiFF}} & 11AM- & 1361895 & 0.3094 & 655704  &  0.3958\\
    &  12AM- & 2136236 & \textbf{0.2814}& 1046258& 0.3542\\
    & Avg. &   & 0.3024 &  & 0.3765\\
    \multirow{3}{*}{SSM}  & 11AM- & 1361895 &0.3178 & 655704 & 0.3624 \\
    &  12AM- & 2136236 &0.2962 & 1046258& \textbf{0.3266}\\
    & Avg. &  &0.3094  & & 0.3478 \\
    \bottomrule
    \end{tabular}
    
    \label{Divergence}
\end{table}

\subsection{Applicability to Downstream Task}

Pedestrian motion prediction is used as a case for evaluating the applicability of our model to downstream tasks. For non-myopic robotic navigation, it’s important that the prediction is made over the entire duration to the destination. In practice, we try to simulate the following scene: a service robot walking down a corridor in a small room, possibly for about 4.8 seconds; and an operating robot walking down a longer corridor in a factory, possibly for about 20 seconds. Therefore, we consider the horizon length over 4.8s and 20s in the rather larger indoor setting of interest, which some current research is lacking at these time spans.

We see our work as macroscopic works, to distinguish it from some microscopic works. Traditional metrics for pedestrian trajectory prediction using microscopic features are Average Displacement Error (ADE) and Final Displacement Error (FDE). ADE is calculated by the mean square error (MSE) over all the displacement in position per person between the prediction and the ground-truth data in the whole trajectory and FDE is calculated at the final endpoint. We use the mean value of the generated distribution to calculate the error and compare it with microscopic methods.

% When there are k trajectory samples generated, the metrics extended to $min_{k}ADE$ and $min_{k}FDE$ with the minimum error selected.

\begin{figure}[htbp]
      \centering
      \includegraphics[width=8cm]{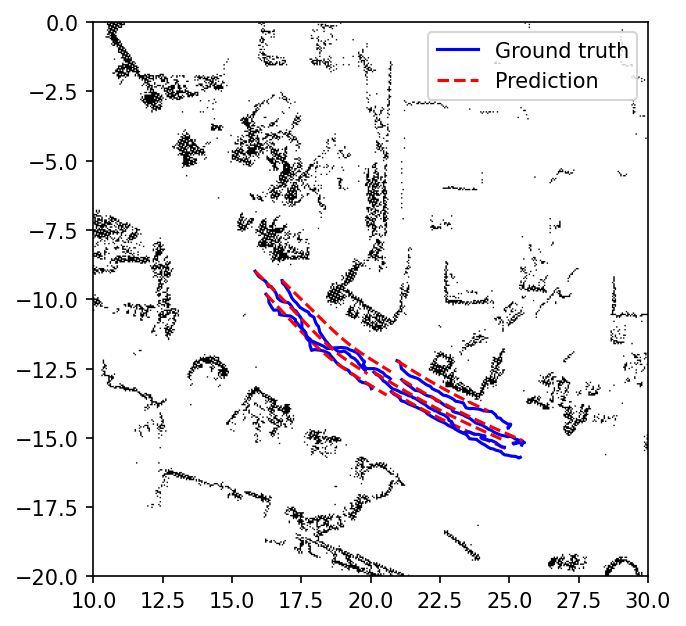}
      \caption{Visualization of the pedestrian motion prediction. The observations in eastern long corridor of ATC dataset is used for training and evaluation. }
      \label{MotionVis}
  \end{figure}

For this task, 0.1s was chosen as a time step and the network observed 50 time steps (5s) in the experiment. We retrained the model with a batch size of 16 in 100 epochs. As shown in Fig. \ref{MotionVis}, the observations in the eastern long corridor of ATC dataset is used for training and evaluation. The results of the applicability in pedestrian motion prediction is shown in Table \ref{motion-prediction}. We compared our method with the state-of-the-art motion prediction algorithm Social GAN (SGAN) \cite{gupta2018social}. SGAN obtains values at every 0.4 seconds and it was designed and trained for 12 time steps (4.8s), when it can get its best performance. As a microscropic method, SGAN generates associated predictions for every pedestrians, but our method has no concept of individual pedestrians for inputs, which is somehow unfair. 

\begin{table}[htbp]
    \centering
    \caption{Results of Pedestrian Motion Prediction}
    \begin{tabular}{lccccc}
    \toprule
   \multirow{2}{*}{Model} & \multirow{2}{*}{Horizon} &  \multicolumn{2}{c}{Sun} & \multicolumn{2}{c}{Wed}\\
   
     &  & ADE(m) & FDE(m) & ADE(m) & FDE(m)\\
    \midrule
    \multirow{2}{*}{SGAN \cite{gupta2018social}} & 4.8s & \textbf{0.6382}& 1.1896 &\textbf{0.6163} & 1.0758 \\
    &  20s &1.8956& 3.6783& 1.9464 & 3.8433\\
    \multirow{2}{*}{SSM}  & 4.8s & 1.1084 &  2.3260 & 1.3941  &2.7436\\
    &  20s & 2.2147 & 3.9748& 1.8420&3.3368\\
    SSM & 4.8s &  0.6824 & \textbf{0.8892} & 0.6761 & \textbf{0.9630}\\
    (one direction)&  20s &\textbf{0.7223} & \textbf{0.9908}& \textbf{0.8828}& \textbf{1.0491}\\
    \bottomrule
    \end{tabular}
    \label{motion-prediction}
\end{table}

In a real-world robotics application,  we can easily determine the direction in which a pedestrian is moving by using sensors. Therefore, we also trained and evaluated our model in only one direction, that is, only consider the orientation $\psi$ in domain $[0, \pi)$ to get a fair comparison. The experimental results demonstrate that our model achieves high accuracy for FDE and long-horizon ADE metrics. However, our model exhibits slight underperformance compared to SGAN for short-horizon ADE. One possible explanation is that our model outputs velocity rather than directly providing trajectory information. As a result, we need to integrate the velocity outputs of each time step to obtain the corresponding position, which is then used as input for the subsequent time step. The above process may introduce additional error, which could contribute to our model's reduced performance. In addition, note that our model was not designed for the motion prediction task, but we can still see that it maintains a certain level of accuracy, which is an encouraging indication of its applicability to downstream tasks.

\section{Conclusion}
In this paper, we presented a method for learning the motion patterns in a changeable environment. The proposed model, leverages a set feature extractor to aggregate spatial information of the input data, a variational autoencoder to encode the spatial information, and a transition module to learn the temporal information. We demonstrated the effectiveness of our method through several experiments, which shows that our model is possible to map the dynamics and be applied in downstream robotic application. 

So far, we have utilized only the temporal and spatial information of dynamic objects, without considering the effects of static and semi-static objects. In the future work, we intend to integrate information from static and semi-static objects to provide better environmental information for robotic applications. Additionally, the position-based velocity fields may vary in different environments. Therefore, introducing semantic information to model motion patterns in diverse environments is another future direction of research.

% A conclusion section is not required. Although a conclusion may review the main points of the paper, do not replicate the abstract as the conclusion. A conclusion might elaborate on the importance of the work or suggest applications and extensions. 

% s\addtolength{\textheight}{-12cm}   % This command serves to balance the column lengths

%%%%%%%%%%%%%%%%%%%%%%%%%%%%%%%%%%%%%%%%%%%%%%%%%%%%%%%%%%%%%%%%%%%%%%%%%%%%%%%%

%%%%%%%%%%%%%%%%%%%%%%%%%%%%%%%%%%%%%%%%%%%%%%%%%%%%%%%%%%%%%%%%%%%%%%%%%%%%%%%%



\section*{Acknowledgements}

% The preferred spelling of the word ÒacknowledgmentÓ in America is without an ÒeÓ after the ÒgÓ. Avoid the stilted expression, ÒOne of us (R. B. G.) thanks . . .Ó  Instead, try ÒR. B. G. thanksÓ. Put sponsor acknowledgments in the unnumbered footnote on the first page.
The authors express their gratitude to Xingyuan Zhang from the Machine Learning Research Lab of Volkswagen Group for his valuable discussions on topics related to this work. 

% His insights and guidance greatly contributed to the completion of this paper.

%%%%%%%%%%%%%%%%%%%%%%%%%%%%%%%%%%%%%%%%%%%%%%%%%%%%%%%%%%%%%%%%%%%%%%%%%%%%%%%%

% References are important to the reader; therefore, each citation must be complete and correct. If at all possible, references should be commonly available publications.

\bibliographystyle{IEEEtran}
\bibliography{IEEEabrv,References}

\end{document}